\documentclass[11pt]{article}
\usepackage{acl}

\usepackage{times}
\usepackage{latexsym}
\usepackage[T1]{fontenc}
\usepackage[utf8]{inputenc}
\usepackage{microtype}
\usepackage{inconsolata}
\usepackage{graphicx}
\usepackage{amsmath,amssymb}
\usepackage{booktabs}
\usepackage{enumitem}
\usepackage{tikz}
\usepackage{multirow}

\title{From Propositional to Perceptual Asymmetry: Extending Frictive Policy Optimization to Asymmetric Partial Information Dialogue}

\author{
  Yifan Zhu
  \and
  Kyeongmin Rim
  \and
  James Pustejovsky \\
  Brandeis University, Waltham, MA, USA \\
  \texttt{\{zhuyifan, krim, jamesp\}@brandeis.edu}
}

\begin{document}
\maketitle

\begin{abstract}

Frictive Policy Optimization \citep[FPO;][]{pustejovsky2025fpo} treats friction in collaborative dialogue -- misalignment, misunderstanding, repair -- 
as an epistemic signal essential to common-ground construction, rather than noise to be minimized.
However, FPO and its implementations  assume shared perceptual contexts, where friction arises from differently interpreted propositions over the same scene, which we define as {\it propositional asymmetry}.
We extend FPO to {\it perceptual asymmetry}, where participants hold asymmetric partial information and the same referring expression yields different denotations depending on
whose information state grounds the reference.
We evaluate this through cross-corpora analysis and LLM probing on referentially asymmetric dialogue tasks, primarily the HCRC MapTask \citep{anderson1991hcrc}.
We find that FPO's friction functional is empirically valid only when evaluated from within each participant's information horizon: different landmark configurations produce qualitatively distinct grounding failure modes, with a small class of ambiguous configurations driving a disproportionate share of misunderstandings through trajectories that appear successful but silently diverge.
The LLM probe confirms that having the ``right perspective'' matters more than having all perspectives: the informed single viewpoint outperforms omniscient access to both participants' contexts.
We propose two annotation refinements: subtype decomposition of pending grounding states and accommodation-aware alignment classification.

\end{abstract}

\section{Introduction}
\label{sec:intro}

Collaborative dialogue relies on common ground, but building that common ground is inherently not frictionless: participants misinterpret, fall out of alignment, and then repair. Rather than treating this friction as noise, Frictive Policy Optimization \citep[FPO;][]{pustejovsky2025fpo} reframes it as an epistemic signal: moments of misunderstanding expose belief-state divergences that policies can exploit for more robust alignment. However, existing implementations of FPO \citep[FAAF;][]{nath-etal-2025-frictional} assume a shared perceptual context where participants reason over the same scene but arrive at different beliefs, which we call \textit{propositional asymmetry}.

Recent work on collaborative construction under epistemic asymmetry points to a qualitatively different setting. In the Distributed Partial Information Puzzle \citep[DPIP;][]{zhu2026distributed}, each participant holds a distinct partial view of the target structure, and successful communication requires reconciling information that was never jointly observed; we call this \textit{perceptual asymmetry}. As formalized by lexical event models \citep{pustejovsky2024lexical}, even identical surface expressions (e.g., ``the red block is on top'') can evoke different event structures depending on which perceptual frame grounds the reference, yielding referential asymmetry as a downstream consequence.


This perceptual divide motivates an extension of FPO to settings where alignment is evaluated relative to each participant’s information horizon, rather than a shared frame. Concretely, we generalize the friction functional to a perspectival formulation with a role-specific decomposition: the speaker incurs friction from within-frame ambiguity at production, while the addressee incurs friction from cross-frame conflicts at interpretation. Under this view, common ground is not presupposed but must be actively constructed by reconciling perceptually disjoint evidence.

Specifically, we make the following contributions. First, through corpus analysis and LLM probing on the HCRC MapTask, we show that even in this controlled cooperative-grounding task, collaborative dialogue exhibits substantial perceptual asymmetry: participants reason over private, non-overlapping perceptual states, making friction perspective-dependent and limiting the reach of shared-scene assumptions. Second, we introduce a perspective-bound evaluation of the friction functional and demonstrate that, compared to an omniscient view, this formulation reveals substantially more frictive states, capturing divergences that would otherwise remain undetected. Finally, motivated by these findings, we propose two targeted refinements to existing grounding-annotation schemes: (i) decomposing unresolved grounding states into finer-grained subtypes aligned with frictive substates, and (ii) distinguishing alignment achieved through negotiation from alignment achieved through silent accommodation by linking grounding trajectories with dialogue-act structure.
Throughout these three contributions, this paper develops the \emph{diagnostic} and conceptual machinery for perspective-bound FPO and demonstrates its empirical purchase on existing dialogue data.

%
\section{Background}
\label{sec:background}

\subsection{LLM Alignment}
\label{sec:bg-alignment}

Mainstream LLM alignment methods, such as RLHF \citep{christiano2017deep}, DPO \citep{rafailov2024direct}, and GRPO \citep{shao2024deepseekmath}, optimize the policy toward a fixed preferred response. In collaborative dialogue this can be counterproductive: an agent that confirms apparent agreement without verifying actual common ground becomes a source of false common ground rather than a safeguard against it. Frictive Policy Optimization \citep[FPO;][]{pustejovsky2025fpo,fpo-theory} proposes that deliberate resistance to immediate task completion via clarification, verification, or challenge, is a rational control action. FPO factors the dialogue policy into two complementary components: an \emph{intervention policy} $\pi_{\text{int}}(a \mid h)$ that decides \emph{whether} (and what type of) friction action $a$ to take given history $h$, and a \emph{generation policy} $\pi_{\text{gen}}(y \mid h, a)$ that realizes the chosen action as an utterance $y$. FPO further defines a structured friction functional $F(h,y) = F^+(h,y) - F^-(h,y)$ that decomposes epistemic risk into independently measurable components. The Frictional Agent Alignment Framework \citep[FAAF;][]{nath-etal-2025-frictional} provides the first empirical implementation, showing that friction-based alignment improves collaborative reasoning on tasks with shared perceptual context. We adopt FPO's friction functional as our theoretical foundation and extend it to settings where FAAF's shared-context assumption does not hold (\S\ref{sec:perspectival-friction}).

\subsection{Grounding in Asymmetric Dialogue}
\label{sec:bg-grounding}
Common ground, the mutual knowledge, beliefs, and assumptions that
participants rely on when coordinating joint actions, is not a static
precondition but a dynamic achievement, constructed incrementally
through the process of \emph{grounding}
\citep{clark1991grounding,clark1996using}.
Grounding succeeds when both participants have sufficient evidence that
their interpretations converge; it fails silently when both proceed as
if convergence holds but their interpretations in fact diverge.


Several task-oriented dialogue corpora study collaborative grounding under information asymmetry: the Weights Task \citep{khebour2024common}, DeliData \citep{karadzhov2023delidata}, the HCRC MapTask \citep{anderson1991hcrc}, OneCommon \citep{udagawa2019natural}, and the Distributed Partial Information Puzzle \citep[DPIP;][]{zhu2026distributed}. These corpora differ in \emph{what} diverges across participants: in \emph{propositional asymmetry}, participants share a scene $S$ but assign different propositional content (Weights Task, DeliData); in \emph{perceptual asymmetry}, each participant observes a different subset of evidence (MapTask, OneCommon, DPIP). We focus on the latter, and within it on \emph{structural} asymmetry: stable, role-bound divergences originating in the organization of the task (e.g., the giver and follower roles in MapTask), which shape the distribution of accessible evidence throughout interaction. Such structurally differentiated access induces downstream perceptual and propositional asymmetries as dialogue unfolds. The relevant point of comparison across corpora is therefore not the surface task domain but the underlying role-conditioned asymmetry structure, which recurs across many collaborative dialogue settings; the perspective-bound formulation applies wherever the evidential history $h$ becomes participant-specific.

\citet{li2026grounded} provide the first annotation scheme tracking
grounding at the level of individual perspectives, separately recording
the speaker's intended and addressee's interpreted landmark for each
reference expression in MapTask (\S~\ref{sec:data-maptask}).
Their key finding is that a small class of multiplicity discrepancies
drives a disproportionate share of misunderstandings, which motivates the
perspective-bound friction taxonomy in
\S~\ref{sec:perspectival-friction}.
They also note an omniscience bias risk: their GPT-5 annotator, with
access to both maps, may underestimate asymmetric knowledge by treating
cross-perspective contradictions as noise, a caveat we operationalize
as a measurable quantity in \S~\ref{sec:perspectival-probe}.

\label{sec:bg-probing}
%

\section{Dataset}
\label{sec:data}

\subsection{HCRC MapTask}
\label{sec:data-maptask}

We draw on two annotation layers over the same 128 HCRC MapTask dialogues \citep{anderson1991hcrc}.
In each dialogue, a giver who holds a printed route guides a follower, who holds a different map of the same area, to reproduce the route.
The two maps share most landmarks but differ in controlled ways: some landmarks appear on one map but not the other (\emph{existence} discrepancies), some share a name but differ in label (\emph{lexical} discrepancies), and a small number appear twice on one map but once on the other (\emph{multiplicity}).

The first layer is the HCRC MapTask NXT annotation release (v2.1, 2011),\footnote{\url{https://groups.inf.ed.ac.uk/maptask/maptasknxt.html}} which packages the original corpus annotations in NXT XML format with dialogue act move labels (\texttt{instruct}, \texttt{check}, \texttt{clarify}, \texttt{acknowledge}, etc.) aligned to word-level timed units.
We use the move labels to identify clarification and verification acts in our accommodation analysis (\S\ref{sec:accommodation}).

\begin{figure}[t]
\centering
\resizebox{\columnwidth}{!}{%
\begin{tikzpicture}[
  utt/.style={draw=black!40, rounded corners=2pt, fill=gray!6, align=left,
              inner sep=3pt, text width=2.8cm, font=\scriptsize},
  re/.style ={draw=blue!45,  rounded corners=2pt, fill=blue!7, align=left,
              inner sep=3pt, text width=3.2cm, font=\scriptsize},
  nxt/.style={font=\scriptsize\ttfamily, fill=orange!18, draw=orange!55,
              rounded corners=1pt, inner sep=1pt},
  cidA/.style={text=red!65!black,    font=\scriptsize\bfseries},
  cidB/.style={text=violet!65!black, font=\scriptsize\bfseries},
  binding/.style={gray!50, very thin},
  chain/.style={->, semithick, draw=red!55!black}
]
\node[font=\scriptsize\itshape, text=black!55, anchor=south] at (1.50, 3.55) {NXT layer};
\node[font=\scriptsize\itshape, text=black!55, anchor=south] at (4.90, 3.55) {GMMT layer};

\node[utt] (u1) at (1.50, 2.90) {U$_1$ Giver \hfill\tikz[baseline=-.5ex]{\node[nxt]{instruct};}\\
                                 \textit{``the church on your left''}};
\node[utt] (u2) at (1.50, 1.55) {U$_2$ Follower \hfill\tikz[baseline=-.5ex]{\node[nxt]{check};}\\
                                 \textit{``the one with the steeple?''}};
\node[utt] (u3) at (1.50, 0.20) {U$_3$ Giver \hfill\tikz[baseline=-.5ex]{\node[nxt]{instruct};}\\
                                 \textit{``then go to a vast meadow''}};

\node[re] (r1) at (4.90, 2.90) {RE$_1$\;{\color{red!65!black}\bfseries [$c_A$]}\;``church''\\
                                g$\!\to\!c_A$\;\;f$\!\to\!c_A$\;\;\textsc{aligned}};
\node[re] (r2) at (4.90, 1.55) {RE$_2$\;{\color{red!65!black}\bfseries [$c_A$]}\;``the one''\\
                                g$\!\to\!c_A$\;\;f$\!\to\!c_A$\;\;\textsc{aligned}};
\node[re] (r3) at (4.90, 0.20) {RE$_3$\;{\color{violet!65!black}\bfseries [$c_B$]}\;``vast meadow''\\
                                g$\!\to\!c_B^{(0)}$\;\;f$\!\to\!c_B^{(1)}$\;\;\textsc{misund.}};

\draw[binding] (u1.east) -- (r1.west);
\draw[binding] (u2.east) -- (r2.west);
\draw[binding] (u3.east) -- (r3.west);

\draw[chain] (r1.east) to[bend left=35]
  node[right=3pt, font=\scriptsize, text=red!55!black] {chain($c_A$)} (r2.east);
\end{tikzpicture}}

\caption{Joined annotation schema for a single dialogue $d$. \textbf{NXT layer}: each utterance carries a dialogue-act tag (orange). \textbf{GMMT layer}: each RE carries a unified concept ID ($c_A$, $c_B$; a superscript marks the multiplicity instance, e.g.\ $c_B^{(0)}$ vs $c_B^{(1)}$), the giver's intended landmark ($g\!\to$), the follower's interpreted landmark ($f\!\to$), and an understanding state in \{\textsc{aligned}, \textsc{pending}, \textsc{misunderstood}\}. REs sharing a concept ID within $d$ form a \emph{reference chain} (red): chain($c_A$) is stable in \textsc{aligned}; chain($c_B$) is a singleton $r_3$ where giver-side multiplicity yields silent misalignment despite a shared name. Joined at \texttt{(dialogue\_id, utterance\_id)}.}
\label{fig:re-chain-schema}
\end{figure}
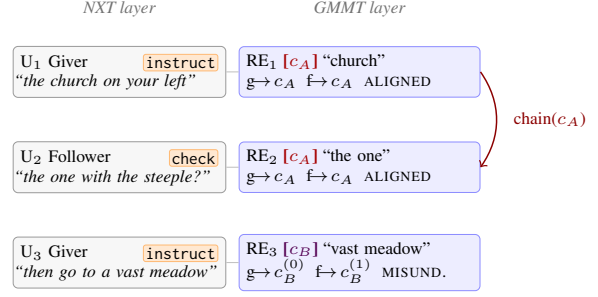

The second layer is the per-perspective annotation by \citet{li2026grounded}, which we refer to as the Grounded Misunderstandings in MapTask dataset (GMMT, our abbreviation for convenience).
GMMT annotates 13,077 reference expressions (REs) across the same 128 dialogues, separately recording the speaker's intended and the addressee's interpreted landmark for each RE, yielding three understanding states: \emph{aligned}, \emph{pending}, and \emph{misunderstood}. REs sharing a concept ID within a dialogue form a \emph{reference chain}, tracking how grounding for a single landmark evolves across the interaction. 
Crucially, GMMT introduces a unified landmark ID scheme that distinguishes multiplicity instances and lexical variants that the original NXT scheme collapses, exposing cases of false common ground invisible under the earlier annotation (Figure~\ref{fig:re-chain-schema}).
This distinction matters: as Table~\ref{tab:misunderstanding-rates} shows, multiplicity discrepancies, though only 6\% of landmarks, account for a 12\% misunderstanding rate---six times the corpus average.

We join the two layers at the (dialogue\_id, utterance\_id) level, giving each GMMT reference expression a list of co-occurring NXT move tags.


\begin{table}[t]
  \centering
  \small
  \begin{tabular}{l rrr}
    \toprule
    \textbf{Discrepancy} & \textbf{REs} & \textbf{Misunderstood} & \textbf{Rate} \\
    \midrule
    Identical     & 8,330 &  18 &  0.2\% \\
    Lexical       &   914 &  13 &  1.4\% \\
    Multiplicity  &   956 & 115 & 12.0\% \\
    Existence     & 2,877 &  93 &  3.2\% \\
    \midrule
    Total         & 13,077 & 239 & 1.8\% \\
    \bottomrule
  \end{tabular}
  \caption{Misunderstanding counts and rates by landmark discrepancy type in the GMMT corpus (replication of \citealp{li2026grounded}, Table~4).}
  \label{tab:misunderstanding-rates}

\end{table}
\subsection{OneCommon}
\label{sec:data-onecommon}

The OneCommon corpus \citep{udagawa2019natural} provides a second referentially asymmetric task: two participants each see a partially overlapping set of dots on a 2D canvas with transformed coordinate frames, and must agree on a shared dot through text dialogue. Among 5,191 annotated dialogues, 1.0\% of markables are flagged \emph{ambiguous} (the expression matches multiple dots in the speaker's view) and 0.3\% \emph{unidentifiable} (the addressee cannot resolve the reference). These tags align with the giver-side and follower-side friction triggers we define in \S\ref{sec:perspectival-friction}. However, OneCommon lacks the perspectivist dual-interpretation structure of GMMT, so we use it as a cross-corpus comparison point in the Discussion (\S\ref{sec:discussion}), not a primary evaluation target.

\section{Perspectival Frictive States}
\label{sec:perspectival-friction}

Existing FPO implementations target tasks with a shared perceptual context (e.g., the Weights Task, where all participants observe the same blocks on the same scale). In that regime, extracting frictive states from the full interaction history $h$ is well-posed, because there is no private perceptual state to miss. When the information structure shifts to an asymmetric partial information setting, such as different maps, different dot configurations, different views of a structure, $h$ is no longer shared: each participant's $h^{(i)}$ includes asymmetric, perspective-specific evidence. As established in~\citet{pustejovsky2024lexical}, epistemic states are shaped by perceived events, and perceptual asymmetries result in epistemic asymmetries. Frictive states therefore should be defined relative to a participant's private perspective.

\label{sec:taxonomy}

\paragraph{Friction functional.} The FPO formulation \citep{fpo-theory} defines a structured friction functional as $F(h, y) = F^{+}(h, y) - F^{-}(h, y)$, where $h$ is the interaction history, $y$ is a candidate response, $F^{+}$ captures productive friction (information gain from clarification or verification), and $F^{-}$ aggregates four epistemic failure modes: uncertainty/miscalibration, contradiction, hazard, and value conflict.

\paragraph{Frictive state.} We define a frictive state as a configuration of the interaction history at turn $t$ in which $F^{-}(h_t, y) > 0$ for at least one plausible response $y$, a configuration where any continuation of the dialogue carries detectable epistemic risk under the friction functional.
This definition makes three commitments. (i)~A frictive state is a property of $h$, not of $y$: it is what $\pi_{\text{int}}$ reads to decide \emph{whether} to intervene, while $\pi_{\text{gen}}$ produces the friction move \emph{given} that decision. (ii)~A frictive state is perspective-relative: when $h$ is participant-specific ($h^{(g)}$, $h^{(f)}$), $F^{-}$ must be evaluated as $F^{-}(h^{(i)}, y)$ for each $i$. (iii)~A frictive state is not necessarily the absence of common ground: both participants may believe they are talking about a parked van (propositionally aligned) while grounding the expression to different physical entities (referentially frictive); the frictive state captures the \emph{risk} this divergence surfaces as task failure.

\paragraph{Failure mode taxonomy.} We adopt the four $F^{-}$ failure modes from the FPO formulation above and instantiate each for referentially asymmetric grounding, specifying what it detects under a perspective-bound information horizon.


\emph{Uncertainty/miscalibration} (within-context referential ambiguity).
FPO measures this substate as miscalibration between expressed confidence and actual reliability. In referential grounding, uncertainty arises when a participant takes up a referring expression but cannot uniquely map it to a landmark: multiple candidates remain plausible, and confidence is low relative to true identifiability. This holds for both giver (unaware of ambiguity) and follower (facing competing candidates).

\emph{Contradiction} (cross-perspective grounding conflict).
FPO flags contradiction as logical inconsistency between a response and the interaction history. In referential grounding, it arises when participants commit to incompatible landmark hypotheses: e.g., the giver says “south of the crane bay,” while the follower’s only viable candidate lies north of it.

\emph{Hazard} (confident miscommitment risk).
FPO uses hazard to capture confident-but-wrong contributions that commit the interaction to costly trajectories. In referential grounding, it arises when a participant confidently acts on a misinterpretation, e.g., the follower routes past the wrong landmark and subsequent turns build on false common ground. The risk is not intent but propagation: high confidence plus undetected misalignment leads error to compound rather than self-correct.

\emph{Value Conflict} (production underspecification).
In FPO, value conflict measures the shift in inferred user intent after committing to $y$; high values indicate “silent intent fixing.” In referential grounding, it arises when an underspecified expression is produced or accepted despite available ambiguity, e.g., the giver says “the vast meadow” with multiple candidates, or the follower commits to one without challenge. In both cases, an interpretation is silently fixed where clarification is warranted.

Table~\ref{tab:diagnostics} reformulates these same four failure modes as diagnostic questions a single participant can ask from their own information horizon $h^{(i)}$, without omniscient access to the partner's evidence.

\begin{table}[t]
  \centering
  \small
  \begin{tabular}{lp{5.2cm}}
    \toprule
    \textbf{Substate} & \textbf{Diagnostic from $h^{(i)}$} \\
    \midrule
    Unc & Can I commit to a referent, or does my evidence admit multiple candidates? \\
    Contr & Given the spatial claims I attribute to my partner, is there a candidate on my map that satisfies both? \\
    Haz & Am I (or my partner) acting confidently on a grounding that I have reason to doubt? \\
    ValConf & Is the expression I am producing (or accepting) uniquely identifying for me, and if not, have I flagged it? \\
    \bottomrule
  \end{tabular}
  \caption{Perspective-bound diagnostic for each frictive substate.}
  \label{tab:diagnostics}

\end{table}

The four components split naturally across the two speaker roles: Value Conflict and Uncertainty fire for a speaker at production, Contradiction and Hazard for an addressee at interpretation. Table~\ref{tab:diagnostics} gives the diagnostic question for each substate, answerable from participant $i$'s information horizon $ h^{(i)}$
 alone, without omniscient access to the partner's private state. We test whether LLMs can replicate these perspectival distinctions in \S~\ref{sec:perspectival-probe}.

We exclude $F^{+}$ (Information Gain) from this taxonomy as $F^{+}$ scores potential future \emph{interventions}, not on past interaction history, treating it as a state type would conflate trigger with response. Its role in connecting state detection to the intervention policy $\pi_{\text{int}}$ is discussed in \S\ref{sec:operationalize}. 

If the multi-component structure is empirically warranted, the four components should be separately anchored in real dialogue evidence. We test this across \S\ref{sec:when-friction}. 




\section{Corpus Analysis}
\label{sec:corpus-analysis}

The taxonomy in \S~\ref{sec:perspectival-friction} makes two testable predictions: (1)~the multi-component structure should be visible in the annotation patterns, different discrepancy types should produce different grounding failure profiles; and (2)~when alignment occurs on difficult landmarks, it should be achieved through friction (negotiation) rather than silent capitulation (accommodation).
We test both using the GMMT and NXT annotations.
The analysis below cross-tabulates three classification axes: landmark \emph{discrepancy type} (Table~\ref{tab:misunderstanding-rates}), GMMT \emph{pending subtype} (Table~\ref{tab:pending-subtypes}), and the four $F^{-}$ \emph{failure modes} from \S\ref{sec:taxonomy} (Table~\ref{tab:diagnostics}). The first two are observational, the third theoretical; relating them is the core empirical move.

\subsection{When is Friction Warranted?}
\label{sec:when-friction}

The GMMT annotation pipeline asks GPT-5~\cite{singh2025openaigpt5card} to answer five binary questions per RE in a fixed order.
The cascade terminates as soon as a question produces a ``pending'' exit; if all five are answered, the system compares each participant's assigned landmark ID to determine \emph{aligned} (same landmark) or \emph{misunderstood} (different landmarks).
GMMT reports results using the three-way status only, but each pending exit point corresponds to a distinct sub-mode already recorded in the released \texttt{extra.subtype} field: \emph{quantificational} (existence queries), \emph{speaker\_query} (speaker-initiated checks),
\emph{unspecified} (no addressee uptake), \emph{unaccommodated} (addressee signals comprehension failure), and \emph{ungrounded} (addressee cannot link the expression to a specific landmark).
We cross-tabulate these existing labels against the discrepancy type of the landmark each RE references; no re-annotation is required.

\begin{table}[t]
  \centering
  \small
  \setlength{\tabcolsep}{3pt}
  \begin{tabular}{@{}l rrrr r@{}}
    \toprule
    \textbf{Subtype} & \textbf{Id.} & \textbf{Lex.} & \textbf{Mul.} & \textbf{Ex.} & \textbf{Tot.} \\
    \midrule
    quantificational & 70.1 & 44.2 & 41.1 & 37.0 & 1604 \\
    speaker\_query   &  4.7 &  8.3 & 10.1 & 14.1 &  363 \\
    unspecified      & 11.0 & 19.2 &  8.0 &  7.5 &  308 \\
    unaccommodated   &  5.5 &  5.8 &  2.7 &  4.3 &  151 \\
    ungrounded       &  8.7 & 22.4 & \textbf{38.2} & \textbf{37.2} &  977 \\
    \midrule
    \# pending       &  949 &  156 &  487 & 1811 & 3403 \\
    \bottomrule
  \end{tabular}
  \caption{Pending subtype profile by discrepancy type (\% within column). Id.~identical, Lex.~lexical, Mul.~multiplicity, Ex.~existence. Tot.~is the row-total count of pending REs per subtype; \# pending~is the column-total count per discrepancy type. Bold highlights the \emph{ungrounded} enrichment under multiplicity and existence; identical landmarks are instead dominated by \emph{quantificational}.}
  \label{tab:pending-subtypes}

\end{table}

Each RE in the corpus references a specific landmark via its concept ID, and each landmark has a \emph{discrepancy type} determined by the map design: \emph{identical} (appears once on both maps), \emph{lexical} (appears on both but with different names), \emph{existence} (appears on one map only), or \emph{multiplicity} (appears twice on one map, once on the other).
Table~\ref{tab:pending-subtypes} cross-tabulates the pending subtype of each RE against the discrepancy type of the landmark it references.
The subtype profiles differ radically across discrepancy types ($\chi^2 = 430.2$, $p < 10^{-84}$, Cram\'{e}r's $V = 0.205$).
Identical-landmark REs are dominated by \emph{quantificational} pending (70.1\%): participants confirm whether shared landmarks exist on both maps, a process that resolves quickly, reflecting $F^+$ information-seeking rather than $F^-$ risk.
Multiplicity REs, by contrast, show a sharply elevated rate of \emph{ungrounded} pending (38.2\% vs.\ 8.7\% for identical; $V = 0.388$): the addressee takes up the reference but cannot confidently link it to a specific instance, because their map contains only one entity where the speaker's contains two, the follower-side Uncertainty diagnostic of Table~\ref{tab:diagnostics}.
Existence REs show a third pattern, enriched for \emph{speaker\_query} (14.1\%) and \emph{ungrounded} (37.2\%): the speaker proactively checks whether the landmark exists on the partner's map ($V = 0.374$), the giver-side ValConf diagnostic.


These results confirm the hypothesis from \S~\ref{sec:taxonomy}: friction-warranting contexts have qualitatively different signatures across discrepancy types, not merely different rates. A friction agent that collapses all pending states into a single ``not yet aligned'' signal would miss the distinction between a routine existence check and a genuine referent-disambiguation failure.

At the chain level, among the 108 multiplicity–discrepancy chains, 37.0\% contain at least one misunderstood referring expression (RE), compared to 0.9\% for identical chains. More strikingly, 38.9\% of multiplicity chains exhibit non-monotonic state trajectories, reaching an aligned state before reverting to pending or misunderstood, whereas this occurs in only 9.9\% of identical chains. This pattern reflects a referent-renegotiation dynamic: alignment on a difficult landmark may be locally achieved yet remains globally unstable, deteriorating as subsequent references reintroduce ambiguity. Such behavior exemplifies the propagation dynamic captured by the Hazard component of $F^{-}$, namely trajectories that appear successful but silently diverge.

This pattern is robust across speaker roles. For giver-produced REs, the misunderstanding rate is 13.9\% on multiplicity landmarks, compared to 0.2\% on identical landmarks; for follower-produced REs, the corresponding rates are 8.4\% and 0.4\%. The higher rate for givers is consistent with the role-based decomposition in \S\ref{sec:taxonomy}: because the giver must choose between multiple candidate instances of a landmark, they are more likely to produce underspecified expressions that trigger misalignment downstream.

\subsection{Accommodation versus Negotiation}
\label{sec:accommodation}

The GMMT ``aligned'' state covers two functionally different cases: \emph{negotiated} alignment, where both participants converge through explicit friction moves (checks, clarifications), and \emph{accommodated} alignment, where one participant silently flips their interpretation without any intervening friction move. We distinguish these by joining the GMMT per-RE interpretation trajectories with NXT dialogue act move tags (\S~\ref{sec:data-maptask}). Operationally, an aligned RE is classified as \emph{accommodated} if the participant's grounded landmark changed from the previous RE in the same chain and no friction-positive move (\texttt{check}, \texttt{query\_yn}, \texttt{query\_w}, \texttt{clarify}, \texttt{align}) occurs within a 3-utterance window.
 
Results in Table~\ref{tab:accommodation} indicate that non-monotonic trajectories are strongly concentrated on multiplicity chains (38.9\% vs.\ 9.9\% for identical; $\chi^2 = 88.6$, $p < 10^{-19}$, $V = 0.231$), confirming that alignment decay is a structural property of referential ambiguity, not a corpus-wide noise floor.
 
\begin{table}[t]
  \centering
  \small
  \begin{tabular}{l rr rr}
    \toprule
    & \multicolumn{2}{c}{\textbf{Aligned REs}} & \multicolumn{2}{c}{\textbf{Chains}} \\
    \cmidrule(lr){2-3} \cmidrule(lr){4-5}
    \textbf{Disc.\ type} & \textbf{Neg.\ \%} & \textbf{Acc.\ \%} & \textbf{Non-mon.\ \%} & \textbf{$n$} \\
    \midrule
    Identical    &  1.2 & 0.1 &  9.9 & 939 \\
    Lexical      &  4.6 & 0.7 & 29.1 &  79 \\
    Multiplicity & 11.9 & 1.1 & 38.9 & 108 \\
    Existence    &  0.5 & 0.0 & 11.9 & 539 \\
    \bottomrule
  \end{tabular}
  \caption{Alignment mode and chain-level non-monotonicity by discrepancy type.
  \emph{Neg.}\ = negotiated (interpretation flip with a friction move in the window);
  \emph{Acc.}\ = accommodated (flip without friction move);
  \emph{Non-mon.}\ = chain contains aligned$\to$pending or aligned$\to$misunderstood.}
  \label{tab:accommodation}

\end{table}
 
The accommodation/negotiation split yields a largely negative result: silent accommodation is rare across the entire corpus (15 cases, 0.2\% of aligned REs; only 4 in multiplicity chains). The dominant pattern is \emph{negotiated} alignment: 11.9\% of multiplicity aligned REs involve a referent flip accompanied by an explicit friction move, vs.\ 1.2\% for identical landmarks ($\chi^2 = 37.7$, $p < 10^{-8}$). Early accommodation does not predict later chain decay (Spearman $\rho = -0.05$, $p = 0.58$, $n = 108$).
 
This negative result is itself informative. MapTask participants \emph{are} effective at deploying friction when interpretations shift: they check, clarify, and renegotiate rather than silently capitulating, and the friction \emph{works} at the RE level. The 39\% chain-level decay rate therefore reflects \emph{new} REs in the same chain resurfacing the underlying ambiguity, not failure of the original repair --- the bottleneck for friction-aware agents is therefore not learning \emph{to} repair, but learning to track that a resolved ambiguity remains structurally available and may resurface across the chain.

\section{LLM Probing: Does Perspective Help Detection?}
\label{sec:perspectival-probe}

\S\ref{sec:corpus-analysis} showed that referential misalignment in GMMT is structured by perspective. In this section, we test whether this perspectival structure is recoverable by an LLM: does restricting the model to a single participant's view improve detection of referential misalignment relative to omniscient access over both maps?

\medskip\noindent\textbf{Data.}\enspace From GMMT we construct a balanced evaluation set of 200 referring expressions, sampled from a pool of 6,098 candidate REs spanning 12,157 utterances across 51 dialogues:
\begin{itemize}[leftmargin=*,nosep,itemsep=2pt]
  \item \textit{Friction-warranted} ($n=100$): REs from the \emph{multiplicity} class with \emph{misunderstood} status, or labeled as \emph{ungrounded} under \emph{pending}. We prioritize human-verified dialogues and complete the set via uniform sampling with a fixed seed.
  \item \textit{Control} ($n=100$): REs with \emph{identical} discrepancy and \emph{aligned} status, from the same dialogues, controlling for lexical and topical variation.
\end{itemize}
The 200-item set is sized for high-precision per-condition comparison rather than corpus-wide generalization; the effect sizes we report ($F_1$ differences $> 0.45$ in every model) are well outside the 95\% confidence intervals at this sample size.

\medskip\noindent\textbf{Models.}\enspace We probe three instruction-tuned LLMs spanning closed and open families, dense and mixture-of-experts architectures, and full- and quantized-precision deployments: \textbf{GPT-5} (closed, frontier-scale; OpenAI API), \textbf{Qwen3.6-35B-A3B} (open, MoE; vLLM, bf16), and \textbf{Llama-4-Scout-17B-16E} (open, MoE; vLLM, w4a16 quantized). Qwen reasoning is disabled to keep outputs short and parseable.

\medskip\noindent\textbf{Conditions.}\enspace Each model is queried under three binary yes/no conditions, each tailored to the epistemic role it instantiates:

\begin{description}[leftmargin=*,nosep,font=\normalfont\textsc]
  \item[C1 -- Omniscient.] Both maps + dialogue prefix. Question: do giver and follower refer to the same physical landmark? Gold positive iff \texttt{misunderstood}.
  \item[C2 -- Giver.] Only the giver's map + dialogue prefix. Question: is the target RE underspecified on the giver's own map? (Production-time ValConf diagnostic.) Gold positive iff the base landmark has multiple instances on the giver's map.
  \item[C3 -- Follower.] Only the follower's map + dialogue prefix. Question: does the giver's spatial description conflict with the follower's map? (Interpretation-time Contr diagnostic.) Gold positive tracks \texttt{misunderstood}.
\end{description}

\begin{table}[t]
\centering
\small
\setlength{\tabcolsep}{2pt}
\renewcommand{\arraystretch}{1.1}
\begin{tabular}{@{}llrrrrrrr@{}}
\toprule
Model & Cond. & Acc. & P & R & $F_1$ & Fr. & Ct. \\
\midrule
\multirow{3}{*}{GPT-5}
  & Omni.  & .684 & .191 & .346 & .247 & .289 & .990 \\
  & Giver  & \textbf{.938} & \textbf{.946} & \textbf{.926} & \textbf{.936} & \textbf{.926} & .949 \\
  & Foll.  & .600 & .184 & .220 & .200 & .239 & .977 \\
\midrule
\multirow{3}{*}{Qwen3.6-35B}
  & Omni.  & .625 & .289 & .512 & .370 & .290 & .960 \\
  & Giver  & \textbf{.955} & \textbf{.960} & \textbf{.950} & \textbf{.955} & \textbf{.950} & .960 \\
  & Foll.  & .725 & .167 & .070 & .098 & .490 & .960 \\
\midrule
\multirow{3}{*}{Llama-4-Scout}
  & Omni.  & .437 & .277 & 1.000 & .434 & .430 & .444 \\
  & Giver  & \textbf{.895} & \textbf{.838} & \textbf{.980} & \textbf{.903} & \textbf{.980} & .810 \\
  & Foll.  & .455 & .183 & .442 & .259 & .280 & .630 \\
\bottomrule
\end{tabular}

\caption{Per-condition probe performance for three LLMs on 200 REs. Conditions (Cond.): \textsc{Omni.}\ shows both maps; \textsc{Giver}/\textsc{Foll.}\ show only that participant's map. P, R, $F_1$ are computed against the per-condition gold label; Fr.\ and Ct.\ report accuracy restricted to the 100-item friction and 100-item control subsets. \textbf{Bold} marks the best value per metric within each model.}
\label{tab:probe}
\end{table}

Evaluated against the human-verified gold from~\citet{li2026grounded}, Table~\ref{tab:probe} shows a consistent pattern across all three models: the omniscient condition performs substantially worse than the giver-perspective condition, \emph{despite having strictly greater access to information}. The giver-versus-omniscient $F_1$ gap is $+0.69$ for GPT-5, $+0.59$ for Qwen3.6, and $+0.47$ for Llama-4-Scout. In every case the gap is driven by the friction subset: control accuracy under the giver condition matches or exceeds that under the omniscient condition, while friction accuracy collapses from $\geq 0.926$ (giver) to $\leq 0.430$ (omniscient).

This is therefore not a uniform degradation but a \emph{specific insensitivity to referential misalignment} that holds across closed and open, dense and MoE, full-precision and quantized models. When the relevant signal --- a single landmark with two instances --- is embedded within the full set of cross-map relations, the model loses the foregrounding that the giver-only condition provides for free. Restricting the model to one participant's view does not strip information; it surfaces the structurally relevant subset.

\paragraph{Error analysis.} Three patterns clarify \emph{how} the omniscient failure manifests at the item level.
(i)~Llama-4-Scout's omniscient $F_1$ is propped up by a near-constant ``yes'' predictor: under C1 it answers yes on $100\%$ of friction items and $55.6\%$ of controls (recall~$=1.000$, false-alarm rate~$=0.541$), yet under C2 the same model recovers an $F_1$ of $0.903$. GPT-5 and Qwen3.6 are more balanced but still positive-biased on friction ($60.5\%$, $72.0\%$).
(ii)~The friction pool splits into $43$ multiplicity-misunderstood items (gold-positive under C1) and $57$ pending-ungrounded items (gold-negative under C1). All three models \emph{conflate} the two under C1 --- Qwen3.6 catches $51\%$ of misunderstood but rules out only $12\%$ of pending-ungrounded; GPT-5 $35\%$ and $26\%$ respectively --- yet classify \emph{both} subtypes at $\geq 88\%$ under C2.
(iii)~$35$ friction items are misclassified by all three models under C1 (pairwise Jaccard $0.50$--$0.64$); only $1$ is misclassified by all three under C2. The omniscient failures are therefore partly item-driven, while the giver-perspective successes are robust across models.
Qualitatively, GPT-5 and Qwen3.6 free-text reasons for shared C1 false negatives \emph{enumerate} the giver's two candidate instances (e.g.\ ``\emph{the shared instance (instance 1)}'') and then conclude alignment. The omniscient prompt thus elicits the structural information but does not license acting on it; the giver prompt foregrounds it as the question.

\section{Perspective-Bound $F^-$ and Repair-Productive Friction}
\label{sec:operationalize}

\S~\ref{sec:perspectival-probe} established that a perspective-restricted view improves an LLM's ability to \emph{detect} referential misalignment. We now test the policy-relevant question: when $F^-$ is computed per-perspective and a frictive state is declared by \emph{disagreement between perspectival readings}, do the detected states correspond to friction the dialogue subsequently \emph{repairs}---i.e., is the perspectival signal not only detectable but productive?

A note on scope: our experiments probe only $F^-$ on history up to the target turn. The $F^+$ component is defined as the expected belief-entropy reduction from a hypothetical intervention, which we do not simulate here. RepairScore (below) serves as a retrospective surrogate: it measures whether flagged frictive states were in fact resolved downstream, operationalizing the $F^+/F^-$ coupling that FPO leaves abstract. A prospective computation of $F^+$, requiring next-turn belief simulation, is deferred to future work.

\paragraph{Perspective-bound contradiction.} Following \S~\ref{sec:taxonomy}, we instantiate the Contradiction component of $F^-$ for referential grounding as
\begin{equation}
\small
\mathrm{Contr}(h^{(i)}, y) = \mathbf{1}\!\left[\mathrm{ref}(y \mid h^{(g)}) \neq \mathrm{ref}(y \mid h^{(f)})\right]
\label{eq:contr}
\end{equation}
where $\mathrm{ref}(y \mid h^{(i)})$ is the landmark to which expression $y$ resolves under participant $i$'s private information horizon $h^{(i)}$. Unlike the omniscient formulation, which evaluates $\mathrm{Contr}(h, y)$ over a global $h$ and is forced to adjudicate cross-map relations, Eq.~\ref{eq:contr} reads each perspective independently and declares friction precisely when the two readings diverge. Rather than asking the model a perspective-aligned question, we \emph{compose} two single-perspective resolutions into a frictive-state detector.

\paragraph{Linking $F^-$ to repair via RepairScore.} Detection alone is insufficient: the FPO framework requires that flagged states correspond to recoverable misalignment, not noise. We use the definition Eq.~\ref{eq:repair-score} of \emph{RepairScore} in \citet{fpo-theory} to evaluate the performance:
\begin{equation}
\small
\mathrm{RepairScore} = \mathbb{E}\!\left[\sum_{t} \mathbf{1}[\mathrm{Contr}(h_t, y_t) > \tau]\, \mathbf{1}_{\mathrm{repair}}(t)\right]
\label{eq:repair-score}
\end{equation}
where $\mathbf{1}_{\mathrm{repair}}(t)$ marks a gold status transition to \texttt{aligned} within the next five REs of the same dialogue. RepairScore operationalizes the $F^+/F^-$ coupling left to future work in the FPO formulation: a frictive-state detector is policy-useful only insofar as the states it flags are ones the dialogue can actually resolve.

\paragraph{Setup.}
We evaluate on 200 REs from 51 dialogues (43 gold contradictions, 13 gold repairs), using both \texttt{Gemma-4-31B}~\cite{gemma_4_2026} and \texttt{GPT-5}. We compare an \emph{omniscient} condition (joint map access) against a \emph{perspectival} condition (independent single-map resolutions with disagreement as the frictive signal).

\begin{table}[t]
\centering
\small
\resizebox{\columnwidth}{!}{%
\begin{tabular}{lcccc}
\toprule
 & \multicolumn{2}{c}{Gemma-4} & \multicolumn{2}{c}{GPT-5} \\
Setting & Omni. & Persp. & Omni. & Persp. \\
\midrule
Repair Count (of 13)      & 5  & \textbf{13} & 12 & \textbf{13} \\
Det.$\rightarrow$Rep. rate & .116 & \textbf{.302} & .279 & \textbf{.302} \\
\bottomrule
\end{tabular}%
}
\caption{Repair-productive friction detection on 200 REs from 51 dialogues.  Repair Count counts repair events preceded by a flag (max 13); Det.→Rep. rate is RepairScore) normalized by $N_{\mathrm{contr}}$.}
\label{tab:repair-estimators}
\vspace{-5mm}
\end{table}


\vspace{-1mm}
 \paragraph{Results.} Table~\ref{tab:repair-estimators} shows that the perspectival setting captures \emph{all 13} gold repair events (under our 5-RE window) for both Gemma-4-31B and GPT-5, whereas the omniscient setting captures 5 and 12 respectively; the detection-to-repair rate converges to $0.302$ under perspective-bound $F^-$ for both models. The omniscient baseline improves with scale, but the gap does not close: even GPT-5 under omniscient access misses one recoverable misalignment, while perspective-bound $F^-$ recovers all under a substantially smaller open-weights model. The perspectival formulation thus does not simply increase detection; it selectively surfaces \emph{the frictions that matter}---those the dialogue is positioned to repair. $F^-$ is empirically valid when evaluated within each participant's information horizon, and the resulting signal is repair-productive in a way that omniscient access, even at frontier scale, does not match.
\vspace{-2mm}
\section{Discussion}
\label{sec:discussion}
\vspace{-2mm}

\paragraph{Annotation refinement 1: subtype decomposition of \emph{pending}.}
\S\ref{sec:when-friction} shows that subtypes of \emph{pending} distribute differently across discrepancy conditions: multiplicity contexts are dominated by \emph{ungrounded} pending (referent-disambiguation failures), whereas identical-landmark contexts are dominated by \emph{quantificational} pending (routine existence checks). We recommend that future perspectivist grounding annotations expose pending subtypes alongside the three-way status, since this information is already in the annotation cascade.
\vspace{-2mm}
\paragraph{Annotation refinement 2: accommodation-aware alignment.}
The \emph{aligned} label similarly conflates \emph{negotiated} alignment (through explicit clarification) and \emph{accommodated} alignment (silent reinterpretation). Although accommodation is rare in MapTask (\S\ref{sec:accommodation}), the conflation hides chain-level non-monotonicity such as the 38.9\% decay rate in multiplicity chains. We recommend distinguishing these modes by linking per-RE trajectories with dialogue-act annotations, as our (\texttt{dialogue\_id}, \texttt{utterance\_id}) join does without re-annotation.

\vspace{-2mm}
\paragraph{Evidence from a second corpus.}
The OneCommon dataset provides complementary evidence that referential asymmetry drives failures of common ground. Its continuously varying asymmetry yields a 56\% selection failure rate, compared to the 1.8\% RE-level misunderstanding rate observed in MapTask. Dialogues containing at least one ambiguous markable also require more turns to complete (5.64 vs.\ 4.53), reinforcing our finding that referential difficulty increases interactional effort. However, surface-level proxies of friction, such as question frequency or hedging, do not predict task success beyond base rates: failed dialogues are in fact longer and contain more hedging than successful ones ($p < 10^{-19}$). This indicates that the quantity of friction is not itself predictive of outcome. Rather, the key determinant is whether friction is \emph{productive} (resolves ambiguity through targeted clarification) or \emph{unproductive} (hedging that fails to repair misalignment), consistent with the $F^{+}/F^{-}$ decomposition. A fully perspectival annotation of OneCommon, applying the grounding cascade separately to each participant’s dot configuration, would operationalize both refinements proposed above and remains ongoing work.

\vspace{-2mm}
\section{Conclusion}
\label{sec:conclusion}
\vspace{-2mm}
Collaborative dialogue is not a shared-context problem: when participants hold private perceptual state, friction lives in the gap between their information horizons. 
We develop the diagnostic and conceptual machinery for perspective-bound FPO, defining frictive states per-perspective with each $F^-$ component computable from one participant's evidence alone.
Three findings follow: (i) multiplicity configurations drive most misunderstandings through trajectories that locally succeed but globally decay; (ii) perspective-bound probing outperforms omniscient access at every scale, a structural rather than capacity-bound limitation; (iii) the same signal tracks the misalignments dialogue actually repairs.
This cuts against a default in agent design—that misalignment is a context-coverage problem solvable by larger windows or stronger models: more context obscured the signal, and a smaller perspectival model beat an omniscient frontier one. 
Wherever agents coordinate under private information—multi-agent LLM systems, human-AI teaming, distributed task handoff—friction-aware systems must reason from a perspective, not over all of them.


\vspace{-4mm}
\paragraph{Future work.} Three directions worth pursuing. (i) A fully perspectival re-annotation of OneCommon \citep{udagawa2019natural}, applying the grounding cascade to each participant's dot configuration separately, would test whether the same $F^-$ diagnostics transfer to continuously varying asymmetry. (ii) Transfer across other structurally asymmetric tasks (DPIP, OneCommon) would evaluate the perspective-bound formulation beyond MapTask. (iii) Existing alignment labels often conflate surface coordination with epistemic alignment, obscuring silent accommodation in which participants appear aligned despite persistent interpretive divergence. A perspectival analysis that tracks Theory-of-Mind belief states could expose these latent divergences and provide a more precise account of epistemic alignment.

\section*{Acknowledgments}
This research was supported by the NSF National AI Institute for Student-AI Teaming (iSAT)
under grants DRL 2019805 and DRL 2454151. The opinions expressed are those of the authors
and do not represent views of the NSF.



\bibliography{custom}

@book{clark1996using,
  author    = {Herbert H. Clark},
  title     = {Using Language},
  publisher = {Cambridge University Press},
  year      = {1996},
}

@incollection{clark1991grounding,
  author    = {Herbert H. Clark and Susan E. Brennan},
  title     = {Grounding in Communication},
  booktitle = {Perspectives on Socially Shared Cognition},
  editor    = {Lauren B. Resnick and John M. Levine and Stephanie D. Teasley},
  pages     = {127--149},
  publisher = {American Psychological Association},
  year      = {1991},
}

@article{anderson1991hcrc,
  author    = {Anne H. Anderson and Miles Bader and Ellen Gurman Bard and Elizabeth Boyle and Gwyneth Doherty and Simon Garrod and Stephen Isard and Jacqueline Kowtko and Jan McAllister and Jim Miller and Catherine Sotillo and Henry S. Thompson and Regina Weinert},
  title     = {The {HCRC} {M}ap {T}ask Corpus},
  journal   = {Language and Speech},
  volume    = {34},
  number    = {4},
  pages     = {351--366},
  year      = {1991},
}

@article{li2026grounded,

  author    = {Nan Li and Albert Gatt and Massimo Poesio},
  title     = {Grounded Misunderstandings in Asymmetric Dialogue: A Perspectivist Annotation Scheme for {MapTask}},
  journal   = {arXiv preprint arXiv:2511.03718},
  year      = {2025},
  note      = {To appear in LREC 2026},
}

@inproceedings{pustejovsky2025fpo,
  author    = {James Pustejovsky and Nikhil Krishnaswamy},
  title     = {Frictive Policy Optimization for {LLM} Agent Interactions},
  booktitle = {Proceedings of the 24th International Conference on Autonomous Agents and Multiagent Systems (AAMAS)},
  year      = {2025},
}

@article{fpo-theory,
  author    = {James Pustejovsky and Nikhil Krishnaswamy},
  title     = {Frictive Policy Optimization for {LLMs}: Epistemic Intervention, Risk-Sensitive Control, and Reflective Alignment},
  journal   = {arXiv preprint arXiv:2604.25136},
  year      = {2026},
}

@article{christiano2017deep,
  author    = {Paul F. Christiano and Jan Leike and Tom Brown and Miljan Martic and Shane Legg and Dario Amodei},
  title     = {Deep Reinforcement Learning from Human Preferences},
  journal   = {Advances in Neural Information Processing Systems},
  volume    = {30},
  year      = {2017},
}

@inproceedings{rafailov2024direct,
  author    = {Rafael Rafailov and Archit Sharma and Eric Mitchell and Christopher D. Manning and Stefano Ermon and Chelsea Finn},
  title     = {Direct Preference Optimization: Your Language Model is Secretly a Reward Model},
  booktitle = {Advances in Neural Information Processing Systems},
  volume    = {36},
  year      = {2024},
}

@article{shao2024deepseekmath,
  author    = {Zhihong Shao and Peiyi Wang and Qihao Zhu and Runxin Xu and Junxiao Song and Xiao Bi and Haowei Zhang and Mingchuan Zhang and Y.K. Li and Y. Wu and others},
  title     = {{DeepSeekMath}: Pushing the Limits of Mathematical Reasoning in Open Language Models},
  journal   = {arXiv preprint arXiv:2402.03300},
  year      = {2024},
}

@article{karadzhov2023delidata,
  author    = {Georgi Karadzhov and Tom Stafford and Andreas Vlachos},
  title     = {{DeliData}: A Dataset for Deliberation in Multi-party Problem Solving},
  journal   = {Proceedings of the ACM on Human-Computer Interaction},
  volume    = {7},
  number    = {CSCW2},
  pages     = {1--25},
  year      = {2023},
}

@inproceedings{khebour2024common,
  author    = {Ibrahim Khalil Khebour and Mariah Bradford and Kenneth Lai and Christopher Tam and Jingxuan Tu and Benjamin A. Ibarra and Nathaniel Blanchard and Nikhil Krishnaswamy and James Pustejovsky},
  title     = {Common Ground Tracking in Multimodal Dialogue},
  booktitle = {Proceedings of the 2024 Joint International Conference on Computational Linguistics, Language Resources and Evaluation (LREC-COLING 2024)},
  pages     = {3587--3602},
  year      = {2024},
}

@inproceedings{udagawa2019natural,
  author    = {Takuma Udagawa and Akiko Aizawa},
  title     = {A Natural Language Corpus of Common Grounding under Continuous and Partially-Observable Context},
  booktitle = {Proceedings of the AAAI Conference on Artificial Intelligence},
  volume    = {33},
  pages     = {7120--7127},
  year      = {2019},
}

@inproceedings{pustejovsky2024lexical,
  title={Lexical event models for multimodal dialogues},
  author={Pustejovsky, James and Zhu, Yifan},
  booktitle={International Conference on Human-Computer Interaction},
  pages={174--192},
  year={2024},
  organization={Springer}
}

@article{zhu2026distributed,
  title={Distributed Partial Information Puzzles: Examining Common Ground Construction Under Epistemic Asymmetry},
  author={Zhu, Yifan and Bradford, Mariah and Lai, Kenneth and Obiso, Timothy and Venkatesha, Videep and Pustejovsky, James and Krishnaswamy, Nikhil},
  journal={arXiv preprint arXiv:2603.05450},
  year={2026}
}

@inproceedings{nath-etal-2025-frictional,
    title = "Frictional Agent Alignment Framework: Slow Down and Don{'}t Break Things",
    author = "Nath, Abhijnan  and
      Graff, Carine  and
      Bachinin, Andrei  and
      Krishnaswamy, Nikhil",
    editor = "Che, Wanxiang  and
      Nabende, Joyce  and
      Shutova, Ekaterina  and
      Pilehvar, Mohammad Taher",
    booktitle = "Proceedings of the 63rd Annual Meeting of the Association for Computational Linguistics (Volume 1: Long Papers)",
    month = jul,
    year = "2025",
    address = "Vienna, Austria",
    publisher = "Association for Computational Linguistics",
    url = "https://aclanthology.org/2025.acl-long.542/",
    doi = "10.18653/v1/2025.acl-long.542",
    pages = "11042--11089",
    ISBN = "979-8-89176-251-0"
}

@misc{gemma_4_2026,
      title={Gemma 4: Open Multimodal Models Based on Gemini 3 Technology}, 
      author={Gemma Team and Google DeepMind},
      year={2026},
      publisher={Google},
      url={https://blog.google/technology/developers/gemma-4-open-models/}
}

@misc{singh2025openaigpt5card,
      title={OpenAI GPT-5 System Card}, 
      author={Aaditya Singh and Adam Fry and Adam Perelman and Adam Tart and Adi Ganesh and Ahmed El-Kishky and Aidan McLaughlin and Aiden Low and AJ Ostrow and Akhila Ananthram and Akshay Nathan and Alan Luo and Alec Helyar and Aleksander Madry and Aleksandr Efremov and Aleksandra Spyra and Alex Baker-Whitcomb and Alex Beutel and Alex Karpenko and Alex Makelov and Alex Neitz and Alex Wei and Alexandra Barr and Alexandre Kirchmeyer and Alexey Ivanov and Alexi Christakis and Alistair Gillespie and Allison Tam and Ally Bennett and Alvin Wan and Alyssa Huang and Amy McDonald Sandjideh and Amy Yang and Ananya Kumar and Andre Saraiva and Andrea Vallone and Andrei Gheorghe and Andres Garcia Garcia and Andrew Braunstein and Andrew Liu and Andrew Schmidt and Andrey Mereskin and Andrey Mishchenko and Andy Applebaum and Andy Rogerson and Ann Rajan and Annie Wei and Anoop Kotha and Anubha Srivastava and Anushree Agrawal and Arun Vijayvergiya and Ashley Tyra and Ashvin Nair and Avi Nayak and Ben Eggers and Bessie Ji and Beth Hoover and Bill Chen and Blair Chen and Boaz Barak and Borys Minaiev and Botao Hao and Bowen Baker and Brad Lightcap and Brandon McKinzie and Brandon Wang and Brendan Quinn and Brian Fioca and Brian Hsu and Brian Yang and Brian Yu and Brian Zhang and Brittany Brenner and Callie Riggins Zetino and Cameron Raymond and Camillo Lugaresi and Carolina Paz and Cary Hudson and Cedric Whitney and Chak Li and Charles Chen and Charlotte Cole and Chelsea Voss and Chen Ding and Chen Shen and Chengdu Huang and Chris Colby and Chris Hallacy and Chris Koch and Chris Lu and Christina Kaplan and Christina Kim and CJ Minott-Henriques and Cliff Frey and Cody Yu and Coley Czarnecki and Colin Reid and Colin Wei and Cory Decareaux and Cristina Scheau and Cyril Zhang and Cyrus Forbes and Da Tang and Dakota Goldberg and Dan Roberts and Dana Palmie and Daniel Kappler and Daniel Levine and Daniel Wright and Dave Leo and David Lin and David Robinson and Declan Grabb and Derek Chen and Derek Lim and Derek Salama and Dibya Bhattacharjee and Dimitris Tsipras and Dinghua Li and Dingli Yu and DJ Strouse and Drew Williams and Dylan Hunn and Ed Bayes and Edwin Arbus and Ekin Akyurek and Elaine Ya Le and Elana Widmann and Eli Yani and Elizabeth Proehl and Enis Sert and Enoch Cheung and Eri Schwartz and Eric Han and Eric Jiang and Eric Mitchell and Eric Sigler and Eric Wallace and Erik Ritter and Erin Kavanaugh and Evan Mays and Evgenii Nikishin and Fangyuan Li and Felipe Petroski Such and Filipe de Avila Belbute Peres and Filippo Raso and Florent Bekerman and Foivos Tsimpourlas and Fotis Chantzis and Francis Song and Francis Zhang and Gaby Raila and Garrett McGrath and Gary Briggs and Gary Yang and Giambattista Parascandolo and Gildas Chabot and Grace Kim and Grace Zhao and Gregory Valiant and Guillaume Leclerc and Hadi Salman and Hanson Wang and Hao Sheng and Haoming Jiang and Haoyu Wang and Haozhun Jin and Harshit Sikchi and Heather Schmidt and Henry Aspegren and Honglin Chen and Huida Qiu and Hunter Lightman and Ian Covert and Ian Kivlichan and Ian Silber and Ian Sohl and Ibrahim Hammoud and Ignasi Clavera and Ikai Lan and Ilge Akkaya and Ilya Kostrikov and Irina Kofman and Isak Etinger and Ishaan Singal and Jackie Hehir and Jacob Huh and Jacqueline Pan and Jake Wilczynski and Jakub Pachocki and James Lee and James Quinn and Jamie Kiros and Janvi Kalra and Jasmyn Samaroo and Jason Wang and Jason Wolfe and Jay Chen and Jay Wang and Jean Harb and Jeffrey Han and Jeffrey Wang and Jennifer Zhao and Jeremy Chen and Jerene Yang and Jerry Tworek and Jesse Chand and Jessica Landon and Jessica Liang and Ji Lin and Jiancheng Liu and Jianfeng Wang and Jie Tang and Jihan Yin and Joanne Jang and Joel Morris and Joey Flynn and Johannes Ferstad and Johannes Heidecke and John Fishbein and John Hallman and Jonah Grant and Jonathan Chien and Jonathan Gordon and Jongsoo Park and Jordan Liss and Jos Kraaijeveld and Joseph Guay and Joseph Mo and Josh Lawson and Josh McGrath and Joshua Vendrow and Joy Jiao and Julian Lee and Julie Steele and Julie Wang and Junhua Mao and Kai Chen and Kai Hayashi and Kai Xiao and Kamyar Salahi and Kan Wu and Karan Sekhri and Karan Sharma and Karan Singhal and Karen Li and Kenny Nguyen and Keren Gu-Lemberg and Kevin King and Kevin Liu and Kevin Stone and Kevin Yu and Kristen Ying and Kristian Georgiev and Kristie Lim and Kushal Tirumala and Kyle Miller and Lama Ahmad and Larry Lv and Laura Clare and Laurance Fauconnet and Lauren Itow and Lauren Yang and Laurentia Romaniuk and Leah Anise and Lee Byron and Leher Pathak and Leon Maksin and Leyan Lo and Leyton Ho and Li Jing and Liang Wu and Liang Xiong and Lien Mamitsuka and Lin Yang and Lindsay McCallum and Lindsey Held and Liz Bourgeois and Logan Engstrom and Lorenz Kuhn and Louis Feuvrier and Lu Zhang and Lucas Switzer and Lukas Kondraciuk and Lukasz Kaiser and Manas Joglekar and Mandeep Singh and Mandip Shah and Manuka Stratta and Marcus Williams and Mark Chen and Mark Sun and Marselus Cayton and Martin Li and Marvin Zhang and Marwan Aljubeh and Matt Nichols and Matthew Haines and Max Schwarzer and Mayank Gupta and Meghan Shah and Melody Huang and Meng Dong and Mengqing Wang and Mia Glaese and Micah Carroll and Michael Lampe and Michael Malek and Michael Sharman and Michael Zhang and Michele Wang and Michelle Pokrass and Mihai Florian and Mikhail Pavlov and Miles Wang and Ming Chen and Mingxuan Wang and Minnia Feng and Mo Bavarian and Molly Lin and Moose Abdool and Mostafa Rohaninejad and Nacho Soto and Natalie Staudacher and Natan LaFontaine and Nathan Marwell and Nelson Liu and Nick Preston and Nick Turley and Nicklas Ansman and Nicole Blades and Nikil Pancha and Nikita Mikhaylin and Niko Felix and Nikunj Handa and Nishant Rai and Nitish Keskar and Noam Brown and Ofir Nachum and Oleg Boiko and Oleg Murk and Olivia Watkins and Oona Gleeson and Pamela Mishkin and Patryk Lesiewicz and Paul Baltescu and Pavel Belov and Peter Zhokhov and Philip Pronin and Phillip Guo and Phoebe Thacker and Qi Liu and Qiming Yuan and Qinghua Liu and Rachel Dias and Rachel Puckett and Rahul Arora and Ravi Teja Mullapudi and Raz Gaon and Reah Miyara and Rennie Song and Rishabh Aggarwal and RJ Marsan and Robel Yemiru and Robert Xiong and Rohan Kshirsagar and Rohan Nuttall and Roman Tsiupa and Ronen Eldan and Rose Wang and Roshan James and Roy Ziv and Rui Shu and Ruslan Nigmatullin and Saachi Jain and Saam Talaie and Sam Altman and Sam Arnesen and Sam Toizer and Sam Toyer and Samuel Miserendino and Sandhini Agarwal and Sarah Yoo and Savannah Heon and Scott Ethersmith and Sean Grove and Sean Taylor and Sebastien Bubeck and Sever Banesiu and Shaokyi Amdo and Shengjia Zhao and Sherwin Wu and Shibani Santurkar and Shiyu Zhao and Shraman Ray Chaudhuri and Shreyas Krishnaswamy and Shuaiqi and Xia and Shuyang Cheng and Shyamal Anadkat and Simón Posada Fishman and Simon Tobin and Siyuan Fu and Somay Jain and Song Mei and Sonya Egoian and Spencer Kim and Spug Golden and SQ Mah and Steph Lin and Stephen Imm and Steve Sharpe and Steve Yadlowsky and Sulman Choudhry and Sungwon Eum and Suvansh Sanjeev and Tabarak Khan and Tal Stramer and Tao Wang and Tao Xin and Tarun Gogineni and Taya Christianson and Ted Sanders and Tejal Patwardhan and Thomas Degry and Thomas Shadwell and Tianfu Fu and Tianshi Gao and Timur Garipov and Tina Sriskandarajah and Toki Sherbakov and Tomer Kaftan and Tomo Hiratsuka and Tongzhou Wang and Tony Song and Tony Zhao and Troy Peterson and Val Kharitonov and Victoria Chernova and Vineet Kosaraju and Vishal Kuo and Vitchyr Pong and Vivek Verma and Vlad Petrov and Wanning Jiang and Weixing Zhang and Wenda Zhou and Wenlei Xie and Wenting Zhan and Wes McCabe and Will DePue and Will Ellsworth and Wulfie Bain and Wyatt Thompson and Xiangning Chen and Xiangyu Qi and Xin Xiang and Xinwei Shi and Yann Dubois and Yaodong Yu and Yara Khakbaz and Yifan Wu and Yilei Qian and Yin Tat Lee and Yinbo Chen and Yizhen Zhang and Yizhong Xiong and Yonglong Tian and Young Cha and Yu Bai and Yu Yang and Yuan Yuan and Yuanzhi Li and Yufeng Zhang and Yuguang Yang and Yujia Jin and Yun Jiang and Yunyun Wang and Yushi Wang and Yutian Liu and Zach Stubenvoll and Zehao Dou and Zheng Wu and Zhigang Wang},
      year={2025},
      eprint={2601.03267},
      archivePrefix={arXiv},
      primaryClass={cs.CL},
      url={https://arxiv.org/abs/2601.03267}, 
}
\appendix

\section{Perspectival Probe: Implementation Details}
\label{app:probe-repro}

We document the construction, model, prompt, parsing, and metric details of the probe reported in \S\ref{sec:perspectival-probe}.

\subsection{Item selection}
The 200-item evaluation set is sampled from the GMMT release~\citep{li2026grounded} under fixed seed $42$, in two strata:
\begin{itemize}[leftmargin=*,nosep]
  \item \textbf{Friction-warranted ($n=100$).} REs whose annotated \texttt{discrepancy\_type} is \texttt{multiplicity} \emph{and} (\texttt{status}=\texttt{misunderstood} \emph{or} \texttt{subtype}=\texttt{ungrounded}). The three human-verified dialogues (\texttt{q1ec2}, \texttt{q1nc3}, \texttt{q1nc7}) are included first; the remainder is filled by uniform sampling without replacement. The resulting subtype composition is $43$ multiplicity-misunderstood and $57$ pending-ungrounded items.
  \item \textbf{Control ($n=100$).} REs with \texttt{discrepancy\_type}=\texttt{identical} and \texttt{status}=\texttt{aligned}, drawn uniformly from the dialogues used by the friction stratum.
\end{itemize}
Gold labels are derived per condition: $g_{\text{C1}}=\mathbb{1}[\text{status}=\text{misund.}]$; $g_{\text{C2}}=\mathbb{1}[\text{giver's map has}\geq 2 \text{ instances of the base concept}]$; $g_{\text{C3}}=g_{\text{C1}}$.

\subsection{Models and decoding}
\begin{itemize}[leftmargin=*,nosep]
  \item \textbf{GPT-5}: OpenAI \texttt{gpt-5} via the Chat Completions API; \texttt{max\_completion\_tokens}=2000, default temperature. One call per (item, condition), with $60$\,s back-off on rate-limit responses. The exact model fingerprint is recorded in every prediction row.
  \item \textbf{Qwen3.6-35B-A3B}: \texttt{Qwen/Qwen3.6-35B-A3B} served with vLLM~$\geq$0.6 in bfloat16, \texttt{tensor\_parallel\_size}=2, \texttt{gpu\_memory\_utilization}=0.85, \texttt{max\_model\_len}=32768. Reasoning is disabled by passing \texttt{enable\_thinking=False} to the chat template and appending a \texttt{/no\_think} suffix to the user message; \texttt{max\_tokens}=1024.
  \item \textbf{Llama-4-Scout-17B-16E}:  served with vLLM using \texttt{quantization="compressed-tensors"} and otherwise identical settings to Qwen; \texttt{max\_tokens}=200.
\end{itemize}
All open-weight decoding is greedy ($T{=}0$, \texttt{top\_p}{=}1.0).

\subsection{Prompts}
Each condition wraps the dialogue prefix (transcript up to and including the target RE's utterance) and the relevant landmark map(s) into a single user-turn prompt that closes with the role-specific question and a strict response template:
\begin{itemize}[leftmargin=*,nosep]
  \item \textbf{C1 (Omniscient).} ``\emph{At this point in the dialogue, are the giver and follower referring to the SAME physical landmark?}'' Both maps rendered as JSON.
  \item \textbf{C2 (Giver).} ``\emph{Is the referring expression \texttt{[expr]} underspecified on YOUR map?}'' Giver's map only.
  \item \textbf{C3 (Follower).} ``\emph{Does the giver's spatial description conflict with YOUR map?}'' Follower's map only.
\end{itemize}
Every prompt closes with the response schema \texttt{\{"answer":"yes"|"no","reason":"..."\}}.

\subsection{Output parsing}
The first balanced JSON object in the raw output is extracted by a brace-counting scan after stripping fenced code-block markers. The \texttt{answer} field is normalized (\texttt{yes}/\texttt{true} $\mapsto$ True, \texttt{no}/\texttt{false} $\mapsto$ False). If JSON extraction fails, a lenient fallback searches the raw text for quoted \texttt{"yes"} / \texttt{"no"} tokens; items still unresolved are dropped from metric computation. The parse-failure rate is below $1\%$ across all nine model$\times$condition cells.

\subsection{Metrics}
For each (model, condition) we report accuracy, precision, recall, and $F_1$ against the gold label fixed by the condition above, together with per-subset accuracy on the friction and control strata.

\end{document}